\documentclass[11pt,a4paper]{article}
\usepackage[hyperref]{main}
\usepackage{times}
\usepackage{latexsym}
\usepackage{graphicx}
\usepackage{caption}
\usepackage{amsmath}
\usepackage{amssymb}
\usepackage{amsfonts}
\usepackage{booktabs}
\usepackage[bottom]{footmisc}
\usepackage{url}
\usepackage{multicol}
\usepackage{qtree}
\graphicspath{{figures/}}

\aclfinalcopy 

\title{	Tree Methods for Hierarchical Classification in Parallel }

\author{Franz A. Heinsen \\
	{\tt franz@glassroom.com} \\ }

\date{September 17, 2022}

\newcommand{\suptag}[1]{^{{\scriptscriptstyle\tt (#1)}}}
\newcommand{\tree}{\suptag{tree}}

\newcommand{\bigO}{\mathcal{O}}

\newcommand{\padval}{\square}
\newcommand{\maskval}{\mu}

\newcommand{\commentstyle}[1]{\scriptsize\sffamily{#1}}
\newcommand{\textcomment}[1]{\text{\commentstyle{#1}}}

\begin{document}
\maketitle

\begin{abstract}
We propose methods that enable efficient hierarchical classification in parallel. Our methods transform a batch of classification scores and labels, corresponding to given nodes in a semantic tree, to scores and labels corresponding to all nodes in the ancestral paths going down the tree to every given node, relying only on tensor operations that execute efficiently on hardware accelerators. We implement our methods and test them on current hardware accelerators with a tree incorporating all English-language synsets in WordNet 3.0, spanning 117,659 classes in 20 levels of depth. We transform batches of scores and labels to their respective ancestral paths, incurring negligible computation and consuming only a fixed 0.04GB of memory over the footprint of data.\footnote{Source code and instructions for replicating our results are online at \href{https://github.com/glassroom/heinsen_routing}{https://github.com/glassroom/heinsen\_tree}.}
\end{abstract}

\section{Introduction}\label{sec:introduction}

\begin{figure*}[t]
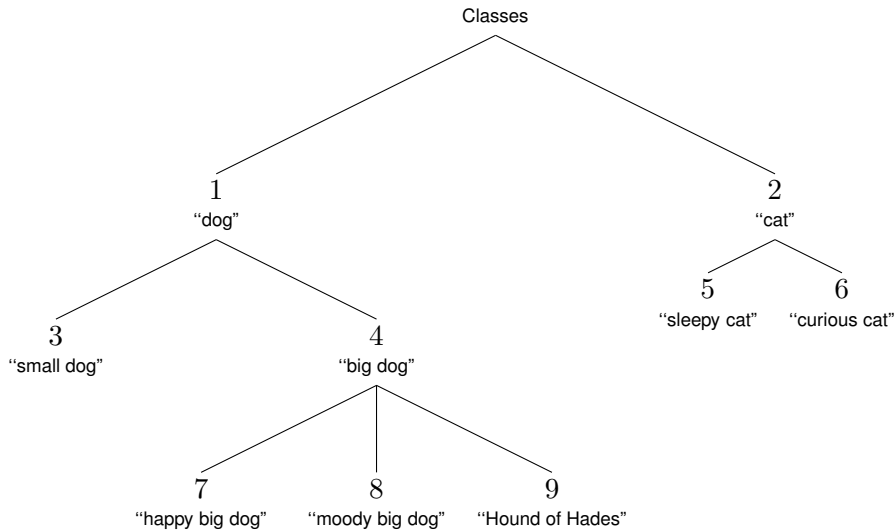

	\vskip 0.1in
	\begin{center}
	\Tree[.{\commentstyle Classes}
		[.{$1$\\[-0.3em] \commentstyle ``dog''}  
			[.{$3$\\[-0.3em] \commentstyle ``small dog''} ]
			[.{$4$\\[-0.3em] \commentstyle ``big dog''} 
				[.{$7$\\[-0.3em] \commentstyle ``happy big dog''} ] 
				[.{$8$\\[-0.3em] \commentstyle ``moody big dog''} ] 
				[.{$9$\\[-0.3em] \commentstyle ``Hound of Hades''} ]
			]
		]
		[.{$2$\\[-0.3em] \commentstyle ``cat''}
			[.{$5$\\[-0.3em] \commentstyle ``sleepy cat''} ]
			[.{$6$\\[-0.3em] \commentstyle ``curious cat''} ]
		]
	]
	\caption{A semantic tree spanning nine nodes in three levels of depth.}
	\label{fig:toy_semantic_tree}
	\end{center}
	\vskip -0.2in
\end{figure*}

Hierarchical classification is a common application of machine learning and artificial intelligence in which models classify data into classes that are nodes of a semantic tree. For illustration, say we want to classify video clips of two kinds of pets, dogs and cats, as either ``dog'' or ``cat'', and further, classify those pets that are dogs as ``small'' or ``big'' and those that are cats as ``sleepy'' or ``curious,'' and further yet, those pets that are dogs that are big as ``happy,'' ``moody,'' or ``Hound of Hades.'' We would train a model---say, a deep neural network pretrained on a large volume of video data, with a new classification head---to learn to classify those video clips into nine classes, each a node of the semantic tree in Figure \ref{fig:toy_semantic_tree}.

A simple approach to hierarchical classification is to train models to classify data into only those classes that are leaf nodes in the tree. In our example (Figure \ref{fig:toy_semantic_tree}), we would train a model to classify data into only six of nine classes: $(3, 5, 6, 7, 8, 9)$. When our model predicts one of those six classes as most probable, say, $8$, we treat its ancestral path, $[1, 4, 8]$, as the hierarchical prediction. The simplicity of this approach makes efficient execution on hardware accelerators trivial, because we can stack a batch of classification scores into a matrix and a batch of labels into a vector, and delegate execution to highly optimized software frameworks for machine learning. However, there is a significant downside: We cannot use any samples labeled with an excluded class to train our model. In our example, if many samples happen to be labeled only at the first level of tree depth, either as ``dog'' ($1$) or ``cat'' ($2$), we would not be able to train our model with those samples, because it does not predict those two labels.

If we want to train our model with samples labeled at all levels of tree depth, we must partition the classification space into subspaces, with at most one class from each ancestral path in each subspace. Classes in the same ancestral path, like ``dog'' ($1$) and ``small dog'' ($3$), cannot be in the same subspace because they are not mutually exclusive. We could partition the nine classes in Figure \ref{fig:toy_semantic_tree} into either three subspaces, $(1, 2)\;(3, 4, 5, 6)\;(7, 8, 9)$, or four subspaces, $(1, 2)\;(3, 4)\;(5, 6)\;(7, 8, 9)$, and train our model to predict a different vector of scores for the classes in each subspace. Alas, there are now two obstacles to efficient execution on hardware accelerators: First, each subspace may have different cardinality, so the vectors of scores may be of different size, and therefore we cannot stack them. Second, if the tree is more than one level deep, the ancestral paths of classes have varying lengths, so the number of ancestral labels per sample varies, and therefore we cannot stack them either.

These two obstacles are not contemplated by the most recent survey of hierarchical classification methods we could find in the literature \cite{10.1007/s10618-010-0175-9}, nor by more recent related work.\footnote{
	We searched for ``hierarchical classification'' on \href{https://arxiv.org}{arXiv} and \href{https://arxiv.org}{GitHub}, and superficially reviewed 134 preprint abstracts and 118 code repositories matching our search term.
} As far as we can discern, the most common approach to hierarchical classification on hardware accelerators is recursively to traverse the tree, transform all scores and labels into score partitions and ancestral labels, respectively, and then either (a) organize data into groups of equal size, stack the data in each group, and process one group at a time, or (b) pad data to common dimensions, stack the padded data, and then process it all at once. The downside to this approach, in both its (a) and (b) variants, is that hardware accelerators are {\em not} optimized for recursively traversing, transforming, grouping, and padding irregularly shaped data. Hardware accelerators are optimized instead for manipulating and transforming data arranged in multidimensional arrays, or tensors.

Here, we propose methods that transform batches of predicted scores and given labels to their corresponding partitioned scores and ancestral paths, relying only on operations algebraically expressible as tensor transformations that execute efficiently on hardware accelerators. We implement the proposed methods and test them on a semantic tree with all English-language synsets of WordNet 3.0 \cite{Fellbaum1998} \cite{10.1145/219717.219748}, spanning 117,659 nodes in 20 levels of depth. Our implementation transforms batches of scores and labels efficiently on recent hardware accelerators, incurring negligible computation and consuming only a fixed modest amount of memory (0.04GB) over the footprint of data, enabling efficient hierarchical classification in parallel.

\subsection{Notation}

In mathematical expressions of tensor transformations, we show all indices as subscript text, implicitly assume broadcasting for any missing indices, perform all operations elementwise, and explicitly show all summations. Superscript text in parenthesis denotes labels. See Table \ref{tab:notation_examples} for examples. We do not use the notation of Linear Algebra, because it cannot handle more than two indices. We do not use Einstein's implicit summation notation either, because it would require the use of operators for raising and lowering indices, adding complexity that is unnecessary for our purposes.

\begin{table}[h]
	\vskip 0.1in
	\small
	\begin{center}
		\begin{tabular}{@{}ll@{}}
			\toprule
			Example                                                              & Implementation in Python \\
			\midrule
			\addlinespace[0.6em]
			$y_{ijk} \longleftarrow x\suptag{1}_{ij} + x\suptag{2}_{jk}$         & {\tt y = x1[:,:,None] + x2} \\
			\addlinespace[0.6em]
			$y_{ijk} \longleftarrow x\suptag{1}_{ij} x\suptag{2}_{jk}$           & {\tt y = x1[:,:,None] * x2} \\
			\addlinespace[0.6em]
			$y_{ik} \longleftarrow \sum_j x\suptag{1}_{ij} x\suptag{2}_{jk}$     & {\tt y = x1 @ x2} \\
			\addlinespace[0.6em]
			$y_{ki} \longleftarrow e^{\sum_j x\suptag{1}_{ij} x\suptag{2}_{jk}}$ & {\tt y = (x1 @ x2).exp().T} \\
			\addlinespace[0.6em]
			$y_{k} \longleftarrow \sum_{ij} x\suptag{1}_{ij} x\suptag{2}_{jk}$   & {\tt y = (x1 @ x2).sum(dim=0)} \\
			\addlinespace[0.6em]
			\bottomrule
		\end{tabular}
	\end{center}
	\caption{\label{tab:notation_examples}Examples of the notation we use, with all-subscript indices, implicit broadcasting, elementwise operations, and explicit summations. In all examples, $x\suptag{1}_{ij} \in \mathbb{R}^{d_1 \times d_2}$ and $x\suptag{2}_{jk} \in \mathbb{R}^{d_2 \times d_3}$.}
\end{table}

\section{Proposed Methods}\label{sec:proposed_methods}

For ease of exposition, we describe our methods as we apply them to the tree in Figure \ref{fig:toy_semantic_tree}. We start by indexing the tree's classes with an index $c$ and its levels of depth with an index $l$:

\begin{equation}
\begin{aligned}
c & = (1, 2, 3, 4, 5, 6, 7, 8, 9) \\
l & = (1, 2, 3).
\end{aligned}
\end{equation}

We partition the classification space by level of depth, satisfying the criteria that at most one class from each ancestral path can be in each subspace. We obtain three subspaces, indexed by $l$:

\begin{equation}
(1, 2) \; (3, 4, 5, 6) \; (7, 8, 9).
\end{equation}

We index samples in a batch with an index $b$. For concreteness, we shall assume that every batch in our example has five samples:

\begin{equation}
b = (1, 2, 3, 4, 5).
\end{equation}

We assume our model outputs one vector with $|c|$ predicted scores for each sample in a batch, and we must partition the vector's elements into $|l|$ vectors for an equal number of subspaces. In our example, there are nine classes, so each vector of predicted scores has nine elements, and we must partition them into three levels of depth.

The methods we propose perform two tasks, expressed algebraically as tensor transformations: First, given a batch of predicted scores $\hat{y}_{bc}$, each an unnormalized log-probability or logit $\in \mathbb{R}$, the task is to partition each row $b$'s scores $c$ by level of depth $l$. Second, given a batch of labels $y_b$, each a positive integer $\in c$, the task is to map them to their corresponding ancestral paths, each a subset of $c$ with up to $|l|$ classes.

\subsection{Encoding and Storing the Tree}

We encode and store the tree once, in advance, in the form of two matrices. One is a matrix of masks $M_{lc}$, whose elements are Boolean values True ($1$) or False ($0$), indicating for each subspace $l$ which classes $c$ are {\em excluded}. In our example:

\begin{equation}\label{eq:M_{lc}}
M_{lc} =
\begin{bmatrix}
0 & 0 & 1 & 1 & 1 & 1 & 1 & 1 & 1 \\
1 & 1 & 0 & 0 & 0 & 0 & 1 & 1 & 1 \\
1 & 1 & 1 & 1 & 1 & 1 & 0 & 0 & 0 \\
\end{bmatrix}.
\end{equation}

The other matrix is a matrix of paths $P_{cl}$, with the ancestral path that ends in each class $c$ as we go down the tree by level $l$. In our example:

\begin{equation}\label{eq:P_{cl}}
P_{cl} =
\begin{bmatrix}
1 & \padval & \padval \\
2 & \padval & \padval \\
1 & 3       & \padval \\
1 & 4       & \padval \\
2 & 5       & \padval \\
2 & 6       & \padval \\
1 & 4       & 7       \\
1 & 4       & 8       \\
1 & 4       & 9       \\
\end{bmatrix}
\begin{array}{l}
\textcomment{// ``dog''} \\
\textcomment{// ``cat''} \\
\textcomment{// ``small dog''} \\
\textcomment{// ``big dog''} \\
\textcomment{// ``sleepy cat''} \\
\textcomment{// ``curious cat''} \\
\textcomment{// ``happy big dog''} \\
\textcomment{// ``moody big dog''} \\
\textcomment{// ``Hound of Hades''} \\
\end{array},
\end{equation}

where $\padval$ is a padding value $\notin c$ (say, $-1$), indicating a path has already reached its class.

Together, these two matrices consume space that is $\bigO \left( (s\suptag{bool} + s\suptag{int}) (|l| \cdot |c|) \right)$, where $s\suptag{bool}$ is the space consumed by each Boolean value in $M_{lc}$, and $s\suptag{int}$ is the space consumed by each integer value in $P_{cl}$. In our example, the matrices consume $(s\suptag{bool} + s\suptag{int}) ( 3 \cdot 9) = 27(s\suptag{bool} + s\suptag{int}$).

\subsection{Partitioning Scores by Level of Depth}

We transform a batch of predicted scores $\hat{y}_{bc}$ into a new tensor $\hat{y}\tree_{blc}$ with scores per sample $b$, partitioned by level of depth $l$, for classes $c$, by applying a masked fill that broadcasts simultaneously over two indices, $b$ and $l$:

\begin{equation}\label{eq:masked_fill}
\hat{y}\tree_{blc} \longleftarrow
\begin{cases}
\maskval,     & M_{lc} = 1 \\
\hat{y}_{bc}, & M_{lc} = 0 \\
\end{cases},
\end{equation}

where $\maskval$, a scalar, is the masking value ({\em e.g.}, the minimum possible score, $-\infty$, or a sentinel value indicating ``not a number,'' or {\tt NaN} for short). That is, we mask those elements of $\hat{y}\tree_{blc}$ whose $lc$ indices coincide with the $lc$ indices where $M_{lc} = 1$, broadcasting over index $b$, and assign $\hat{y}_{bc}$ to the elements in every slice of $\hat{y}\tree_{blc}$ indexed by $bc$ that are not masked, broadcasting over index $l$.

When implemented with a software framework for machine learning, the masked fill \eqref{eq:masked_fill} assigns to each element of $\hat{y}\tree_{blc}$ either the score masking value $\maskval$ or a value in $\hat{y}_{bc}$. Such assignments incur negligible computation and consume only the incremental memory occupied by $\hat{y}\tree_{blc}$.

In our example, the model's predicted scores $\hat{y}_{bc}$ for a batch of five samples are:

\begin{equation}
\hat{y}_{bc} =
{
	\begingroup
	\setlength\arraycolsep{0.1em}
	\begin{bmatrix}
	\hat{y}_{11} & \hat{y}_{12} & \hat{y}_{13} & \hat{y}_{14} & \hat{y}_{15} & \hat{y}_{16} & \hat{y}_{17} & \hat{y}_{18} & \hat{y}_{19} \\	
	\hat{y}_{21} & \hat{y}_{22} & \hat{y}_{23} & \hat{y}_{24} & \hat{y}_{25} & \hat{y}_{26} & \hat{y}_{27} & \hat{y}_{28} & \hat{y}_{29} \\
	\hat{y}_{31} & \hat{y}_{32} & \hat{y}_{33} & \hat{y}_{34} & \hat{y}_{35} & \hat{y}_{36} & \hat{y}_{37} & \hat{y}_{38} & \hat{y}_{39} \\
	\hat{y}_{41} & \hat{y}_{42} & \hat{y}_{43} & \hat{y}_{44} & \hat{y}_{45} & \hat{y}_{46} & \hat{y}_{47} & \hat{y}_{48} & \hat{y}_{49} \\
	\hat{y}_{51} & \hat{y}_{52} & \hat{y}_{53} & \hat{y}_{54} & \hat{y}_{55} & \hat{y}_{56} & \hat{y}_{57} & \hat{y}_{58} & \hat{y}_{59} \\
	\end{bmatrix}
	\endgroup
},
\end{equation}

which the masked fill \eqref{eq:masked_fill} transforms into a tensor $\hat{y}\tree_{blc}$ with $5 \times 3 \times 9$ elements, consisting of the predicted scores for the five samples in the batch, masked at three levels of depth, for nine possible classes. We show the three-dimensional tensor here as five stacked slices of $3 \times 9$ elements:

\begin{equation}\label{eq:transformed_scores}
\hat{y}\tree_{blc} =
{\small
	\begingroup
	\setlength\arraycolsep{0.08em}
	\begin{bmatrix}
		\\[-0.6em]
		\begin{bmatrix}
		\hat{y}_{11} & \hat{y}_{12} & \maskval & \maskval & \maskval & \maskval & \maskval & \maskval & \maskval \\
		\maskval & \maskval & \hat{y}_{13} & \hat{y}_{14} & \hat{y}_{15} & \hat{y}_{16} & \maskval & \maskval & \maskval \\
		\maskval & \maskval & \maskval & \maskval & \maskval & \maskval & \hat{y}_{17} & \hat{y}_{18} & \hat{y}_{19} \\
		\end{bmatrix} \\
		\\[-0.6em]
		\begin{bmatrix}
		\hat{y}_{21} & \hat{y}_{22} & \maskval & \maskval & \maskval & \maskval & \maskval & \maskval & \maskval \\
		\maskval & \maskval & \hat{y}_{23} & \hat{y}_{24} & \hat{y}_{25} & \hat{y}_{26} & \maskval & \maskval & \maskval \\
		\maskval & \maskval & \maskval & \maskval & \maskval & \maskval & \hat{y}_{27} & \hat{y}_{28} & \hat{y}_{29} \\
		\end{bmatrix} \\
		\\[-0.6em]
		\begin{bmatrix}
		\hat{y}_{31} & \hat{y}_{32} & \maskval & \maskval & \maskval & \maskval & \maskval & \maskval & \maskval \\
		\maskval & \maskval & \hat{y}_{33} & \hat{y}_{34} & \hat{y}_{35} & \hat{y}_{36} & \maskval & \maskval & \maskval \\
		\maskval & \maskval & \maskval & \maskval & \maskval & \maskval & \hat{y}_{37} & \hat{y}_{38} & \hat{y}_{39} \\
		\end{bmatrix} \\
		\\[-0.6em]
		\begin{bmatrix}
		\hat{y}_{41} & \hat{y}_{42} & \maskval & \maskval & \maskval & \maskval & \maskval & \maskval & \maskval \\
		\maskval & \maskval & \hat{y}_{43} & \hat{y}_{44} & \hat{y}_{45} & \hat{y}_{46} & \maskval & \maskval & \maskval \\
		\maskval & \maskval & \maskval & \maskval & \maskval & \maskval & \hat{y}_{47} & \hat{y}_{48} & \hat{y}_{49} \\
		\end{bmatrix} \\
		\\[-0.6em]
		\begin{bmatrix}
		\hat{y}_{51} & \hat{y}_{52} & \maskval & \maskval & \maskval & \maskval & \maskval & \maskval & \maskval \\
		\maskval & \maskval & \hat{y}_{53} & \hat{y}_{54} & \hat{y}_{55} & \hat{y}_{56} & \maskval & \maskval & \maskval \\
		\maskval & \maskval & \maskval & \maskval & \maskval & \maskval & \hat{y}_{57} & \hat{y}_{58} & \hat{y}_{59} \\
		\end{bmatrix} \\
		\\[-0.6em]
	\end{bmatrix}.
	\endgroup
}
\end{equation}

\subsection{Mapping Labels to Ancestral Paths}

We map a batch of given labels $y_b$ to a matrix $y\tree_{bl}$, consisting of ancestral labels for each sample $b$ as we go down levels of depth $l$, by referencing the rows of $P_{cl}$ that are indexed by $y_b$ itself:

\begin{equation}\label{eq:y_tree}
y\tree_{bl} \longleftarrow 
\begin{bmatrix}
\vdots \\
P_{y^*_b l} \\
\vdots \\
\end{bmatrix}
\end{equation}

where $P_{y^*_b l}$ denotes a pointer to rows $y_b$ of $P_{cl}$. That is, we use each class as the index to its own ancestral path in the tree, stored in $P_{cl}$. Recall that every element of $y_b$ is a class $\in c$.

When implemented, the referencing of $P_{cl}$'s rows by their index (that is, via preexisting pointers) incurs negligible computation and consumes only the incremental memory occupied by $y\tree_{bl}$.

In our example, if the given labels $y_b$ are, say,

\begin{equation}
y_b =
\begin{bmatrix} 4 \\ 7 \\ 2 \\ 6 \\ 3 \end{bmatrix}
\begin{array}{l}
\textcomment{// ``big dog''} \\
\textcomment{// ``happy big dog''} \\
\textcomment{// ``cat''} \\
\textcomment{// ``curious cat''} \\
\textcomment{// ``small dog''} \\
\end{array},
\end{equation}

the corresponding ancestral paths $y\tree_{bl}$ are:

\begin{equation}\label{eq:transformed_labels}
y\tree_{bl} =
\begin{bmatrix}
P_{y^*_1 l} \\
P_{y^*_2 l} \\
P_{y^*_3 l} \\
P_{y^*_4 l} \\
P_{y^*_5 l} \\
\end{bmatrix}
=
\begin{bmatrix}
1 & 4       & \padval  \\
1 & 4       & 7       \\
2 & \padval & \padval \\
2 & 6       & \padval \\
1 & 3       & \padval \\
\end{bmatrix}.
\end{equation}

\section{Hierarchical Classification in Parallel}

\subsection{Training}

For training a model, we flatten indices $bl$ in the tree-dimensional tensor $\hat{y}\tree_{blc}$ and the matrix $y\tree_{bl}$ to obtain, respectively, a matrix of masked scores and a vector of ancestral labels and padding values. In our example, if we flatten indices $bl$ as described in both \eqref{eq:transformed_scores} and \eqref{eq:transformed_labels}, we obtain:

\begin{equation}
{\small
	\begingroup
	\setlength\arraycolsep{0.1em}
	\begin{bmatrix}
	\hat{y}_{11} & \hat{y}_{12} & \maskval & \maskval & \maskval & \maskval & \maskval & \maskval & \maskval \\
	\maskval & \maskval & \hat{y}_{13} & \hat{y}_{14} & \hat{y}_{15} & \hat{y}_{16} & \maskval & \maskval & \maskval \\
	\maskval & \maskval & \maskval & \maskval & \maskval & \maskval & \hat{y}_{17} & \hat{y}_{18} & \hat{y}_{19} \\
	
	\hat{y}_{21} & \hat{y}_{22} & \maskval & \maskval & \maskval & \maskval & \maskval & \maskval & \maskval \\
	\maskval & \maskval & \hat{y}_{23} & \hat{y}_{24} & \hat{y}_{25} & \hat{y}_{26} & \maskval & \maskval & \maskval \\
	\maskval & \maskval & \maskval & \maskval & \maskval & \maskval & \hat{y}_{27} & \hat{y}_{28} & \hat{y}_{29} \\
	
	\hat{y}_{31} & \hat{y}_{32} & \maskval & \maskval & \maskval & \maskval & \maskval & \maskval & \maskval \\
	\maskval & \maskval & \hat{y}_{33} & \hat{y}_{34} & \hat{y}_{35} & \hat{y}_{36} & \maskval & \maskval & \maskval \\
	\maskval & \maskval & \maskval & \maskval & \maskval & \maskval & \hat{y}_{37} & \hat{y}_{38} & \hat{y}_{39} \\
	
	\hat{y}_{41} & \hat{y}_{42} & \maskval & \maskval & \maskval & \maskval & \maskval & \maskval & \maskval \\
	\maskval & \maskval & \hat{y}_{43} & \hat{y}_{44} & \hat{y}_{45} & \hat{y}_{46} & \maskval & \maskval & \maskval \\
	\maskval & \maskval & \maskval & \maskval & \maskval & \maskval & \hat{y}_{47} & \hat{y}_{48} & \hat{y}_{49} \\
	
	\hat{y}_{51} & \hat{y}_{52} & \maskval & \maskval & \maskval & \maskval & \maskval & \maskval & \maskval \\
	\maskval & \maskval & \hat{y}_{53} & \hat{y}_{54} & \hat{y}_{55} & \hat{y}_{56} & \maskval & \maskval & \maskval \\
	\maskval & \maskval & \maskval & \maskval & \maskval & \maskval & \hat{y}_{57} & \hat{y}_{58} & \hat{y}_{59} \\
	\end{bmatrix}
	\endgroup
}
{\small
	\begin{bmatrix}
	1 \\ 4       \\ \padval \\
	1 \\ 4       \\ 7       \\
	2 \\ \padval \\ \padval \\
	2 \\ 6       \\ \padval \\
	1 \\ 3       \\ \padval \\
	\end{bmatrix}
}.
\end{equation}

We remove all rows in which the vector has a padding value to obtain a smaller matrix of scores and a vector of labels corresponding to the ancestral paths going down the tree to every given node:

\begin{equation}
{\small
	\begingroup
	\setlength\arraycolsep{0.1em}
	\begin{bmatrix}
	\hat{y}_{11} & \hat{y}_{12} & \maskval & \maskval & \maskval & \maskval & \maskval & \maskval & \maskval \\
	\maskval & \maskval & \hat{y}_{13} & \hat{y}_{14} & \hat{y}_{15} & \hat{y}_{16} & \maskval & \maskval & \maskval \\
	
	\hat{y}_{21} & \hat{y}_{22} & \maskval & \maskval & \maskval & \maskval & \maskval & \maskval & \maskval \\
	\maskval & \maskval & \hat{y}_{23} & \hat{y}_{24} & \hat{y}_{25} & \hat{y}_{26} & \maskval & \maskval & \maskval \\
	\maskval & \maskval & \maskval & \maskval & \maskval & \maskval & \hat{y}_{27} & \hat{y}_{28} & \hat{y}_{29} \\
	
	\hat{y}_{31} & \hat{y}_{32} & \maskval & \maskval & \maskval & \maskval & \maskval & \maskval & \maskval \\
	
	\hat{y}_{41} & \hat{y}_{42} & \maskval & \maskval & \maskval & \maskval & \maskval & \maskval & \maskval \\
	\maskval & \maskval & \hat{y}_{43} & \hat{y}_{44} & \hat{y}_{45} & \hat{y}_{46} & \maskval & \maskval & \maskval \\
	
	\hat{y}_{51} & \hat{y}_{52} & \maskval & \maskval & \maskval & \maskval & \maskval & \maskval & \maskval \\
	\maskval & \maskval & \hat{y}_{53} & \hat{y}_{54} & \hat{y}_{55} & \hat{y}_{56} & \maskval & \maskval & \maskval \\
	\end{bmatrix}
	\endgroup
	\begin{bmatrix}
	1 \\ 4       \\
	1 \\ 4       \\ 7       \\
	2 \\ 
	2 \\ 6       \\
	1 \\ 3       \\
	\end{bmatrix}
},
\end{equation}

enabling us to compute a classification loss ({\em e.g}., cross-entropy) at all levels of depth in parallel. The flattening of two indices and the filtering of rows by padding values incur negligible computation and consume no incremental memory.

\subsection{Inference}

At inference, we can obtain predicted probability distributions $\hat{p}\tree_{blc}$ for every sample $b$ at each level of depth $l$ over classes $c$, with a single Softmax function normalizing over index $c$:

\begin{equation}
\hat{p}\tree_{blc} = \frac{ \exp (\hat{y}\tree_{blc}) }{ \sum_c \exp (\hat{y}\tree_{blc}) },
\end{equation}

which, when implemented, is efficiently executed as long as elements with masking value $\mu$ ({\em e.g.}, $-\infty$, {\tt NaN}) are ignored on-device.

If we naively use the class with the highest predicted probability at each level of depth as our prediction, we may obtain nonsensical predictions like $[1, 6, 7]$ (a ``dog'' that is a ``curious cat'' that is a ``happy big dog''). To prevent such nonsense, we can restrict the space of allowed predictions to only valid paths that exist in $P_{cl}$ \eqref{eq:P_{cl}}. We have at our disposal multiple well-known techniques for finding the paths in $P_{cl}$ that most closely match the naively predicted paths, including beam search over the top $k$ paths of $P_{cl}$ with highest joint predicted probability in $\hat{p}\tree_{blc}$ at each level of depth, or selection of the top $k$ paths in $P_{cl}$ with the smallest Levenshtein distance to the naively predicted paths obtained from $\hat{p}\tree_{blc}$.

\section{Implementation and Test}

We implement the proposed methods as a composable open-source library, and test it on a semantic tree of all synsets in WordNet 3.0. Every child-to-parent connection represents a hyponym-hypernym (child ``is a specific instance of'' parent) relationship between two synsets. Some synsets have two or more possible ancestral paths.\footnote{
	For example, the synset for the most common meaning of ``dog'' has two possible ancestral paths in WordNet 3.0: (a) {\scriptsize \sffamily [``entity'', ``physical entity'', ``object'', ``whole'', ``living thing'', ``organism'', ``animal'', ``chordate'', ``vertebrate'', ``mammal'', ``placental'', ``carnivore'', ``canine'', ``dog'']}, and (b) {\scriptsize \sffamily  [``entity'', ``physical entity'', ``object'', ``whole'', ``living thing'', ``organism'', ``animal'', ``domestic animal'', ``dog'']}.
} In those cases, we incorporate only one ancestral path in the tree. The tree has 117,659 synsets, each a class, distributed over 20 levels of depth.

We test our implementation on recent Nvidia GPUs, with batches of 100 samples, each with a vector of 117,659 predicted scores and a given label, and transform them on-device into all corresponding partitioned scores and ancestral paths. Computation is negligible, as expected, as it consists entirely of assignments that are executed efficiently on-device. Average execution time is negligible as well: On a single recent Nvidia GPU, it takes on the order of $1 ms$ to transform all score vectors in a batch, and on the order of $10 ns$ to transform all labels in a batch, to their respective ancestral paths in the WordNet 3.0 tree.

Table \ref{tab:memory_consumption} shows memory consumption on recent Nvidia GPUs using default data types (e.g., float32 instead of float16). Our software library consumes only a fixed 40MB of GPU memory, including the space it uses to store the tree in advance as matrices $M_{lc}$ and $P_{cl}$, with $20 \times 117659$ and $117659 \times 20$ elements, respectively. The only additional memory consumed is occupied by given data ($\hat{y}_{bc}$, $y_b$) and transformed data ($\hat{y}\tree_{blc}$,  $y\tree_{bl}$).

\begin{table}[h]
	\small
	\begin{center}
		\begin{tabular}{lccr}
			\toprule
			\bf Objects                  & \bf Shape                     & \bf Type    & \bf MB \\
			\midrule
			\addlinespace[0.6em]
			\em Given data & & & \\
			Scores $\hat{y}_{bc}$        & $100 \times 117659$           & float32     &  44.9 \\
			Labels $y_b$                 & $100$                         & int64       &   0.0 \\
			\addlinespace[0.6em]
			\em Transformed & & & \\
			Scores $\hat{y}\tree_{blc}$  & $100 \times 20 \times 117659$ & float32     & 897.7 \\
			Labels $y\tree_{bl}$         & $100 \times 20$               & int64       &   0.0 \\
			\midrule
			Total for data               & --                            & --          & 942.6 \\	
			Our software*                 & --                            & --          &  40.0 \\	
			\bottomrule
		\end{tabular} \\
		* Includes memory consumed by matrices $M_{lc}$ and $P_{cl}$.
	\end{center}
	\caption{\label{tab:memory_consumption}Memory consumption on recent Nvidia GPUs for transforming 100 score vectors and given labels to a tree spanning 117,659 classes in 20 levels of depth, using default data types.}
\end{table}


\bibliography{main}
\bibliographystyle{main}

\end{document}